\documentclass[10pt,twocolumn,letterpaper]{article}

\usepackage[pagenumbers]{cvpr} 

\usepackage{wrapfig}
\usepackage{url}
\usepackage[table]{xcolor}
\usepackage{booktabs}
\usepackage{amsmath, amssymb, amsthm, bm}
\usepackage{amsfonts}

\usepackage{multirow}
\usepackage{mathtools}
\usepackage{algorithm}

\usepackage{algorithmicx}
\usepackage{adjustbox}
\usepackage{algpseudocode}
\usepackage{soul}
\usepackage{wrapfig}
\usepackage{breqn}
\newcommand{\conf}[1]{{\color{gray}{\tiny{[{#1}]\!}}}}

\newcommand{\cmark}{\checkmark}
\newcommand{\xmark}{\(\times\)}

\theoremstyle{plain}

\theoremstyle{remark}

\theoremstyle{definition}

\DeclareMathOperator*{\argmin}{argmin}

\usepackage{pifont}

\definecolor{bblue}{RGB}{0,30,95}
\definecolor{rred}{RGB}{190,0,0}
\definecolor{mygray}{gray}{0.1}
\definecolor{ggray}{RGB}{127,127,127}
\definecolor{rowgray}{RGB}{245,245,245}

\definecolor{cvprblue}{rgb}{0.21,0.49,0.74}
\usepackage[breaklinks,hidelinks]{hyperref}

\def\paperID{1349} 
\def\confName{CVPR}
\def\confYear{2026}

\title{Revisiting Prototype Rehearsal for Exemplar-Free Continual Learning: Manifold-Aware Boundary Sampling with Adaptive Class\mbox{-}Balanced Loss}

\author{
Hongye Xu\\
Chester F. Carlson Center for Imaging Science\\
Rochester Institute of Technology\\
{\tt\small hx5239@rit.edu}
\and
Bartosz Krawczyk\\
Chester F. Carlson Center for Imaging Science\\
Rochester Institute of Technology\\
{\tt\small bartosz.krawczyk@rit.edu}
}

\begin{document}
\maketitle
\begin{abstract}

Exemplar-free class-incremental learning (EFCIL) aims to acquire new classes over time without storing raw data. Historically, prototype rehearsal—sampling around stored class prototypes and mixing them with current-task data—has been a popular strategy to reduce catastrophic forgetting. However, recent drift-compensation methods that explicitly realign prototypes in the evolving feature space consistently outperform prototype-based rehearsal, raising the question of whether rehearsal itself is fundamentally limited. We argue that the performance gap stems not from the idea of prototype rehearsal per se, but from how it is typically instantiated: existing approaches (i) treat prototypes as isolated class summaries that ignore information from nearby enemy classes, and (ii) fail to correct the emerging class imbalance between a handful of synthetic old-class samples and hundreds of real instances from newly introduced classes. Building on this hypothesis, we revisit prototype rehearsal and propose a manifold-aware variant that restores its competitiveness in EFCIL. First, we introduce Constrained Expansive Over-Sampling, which interpolates each old-class prototype toward its nearest enemy features from new classes, generating boundary-aware rehearsal samples that better follow the underlying data manifold while preserving inter-class separation. Second, we design an Adaptive Class-Balanced loss that performs time-based class weighting, amplifying gradients from older prototypes when they are most informative and gradually annealing their influence as richer supervision from later tasks accumulates. Together, these components turn prototype rehearsal into a drift-resilient, imbalance-aware mechanism that closes—and often reverses—the gap to recent drift-compensation methods, achieving state-of-the-art performance across multiple EFCIL benchmarks. The code is available
at \textcolor{blue}{\url{https://github.com/HXuSz11/ACB_CEOS_CVPR2026_Findings}}.
\end{abstract}



\vspace{-6mm}
\section{Introduction}
\label{in}
\begin{figure}[t]
  \centering
  \includegraphics[width=\linewidth]{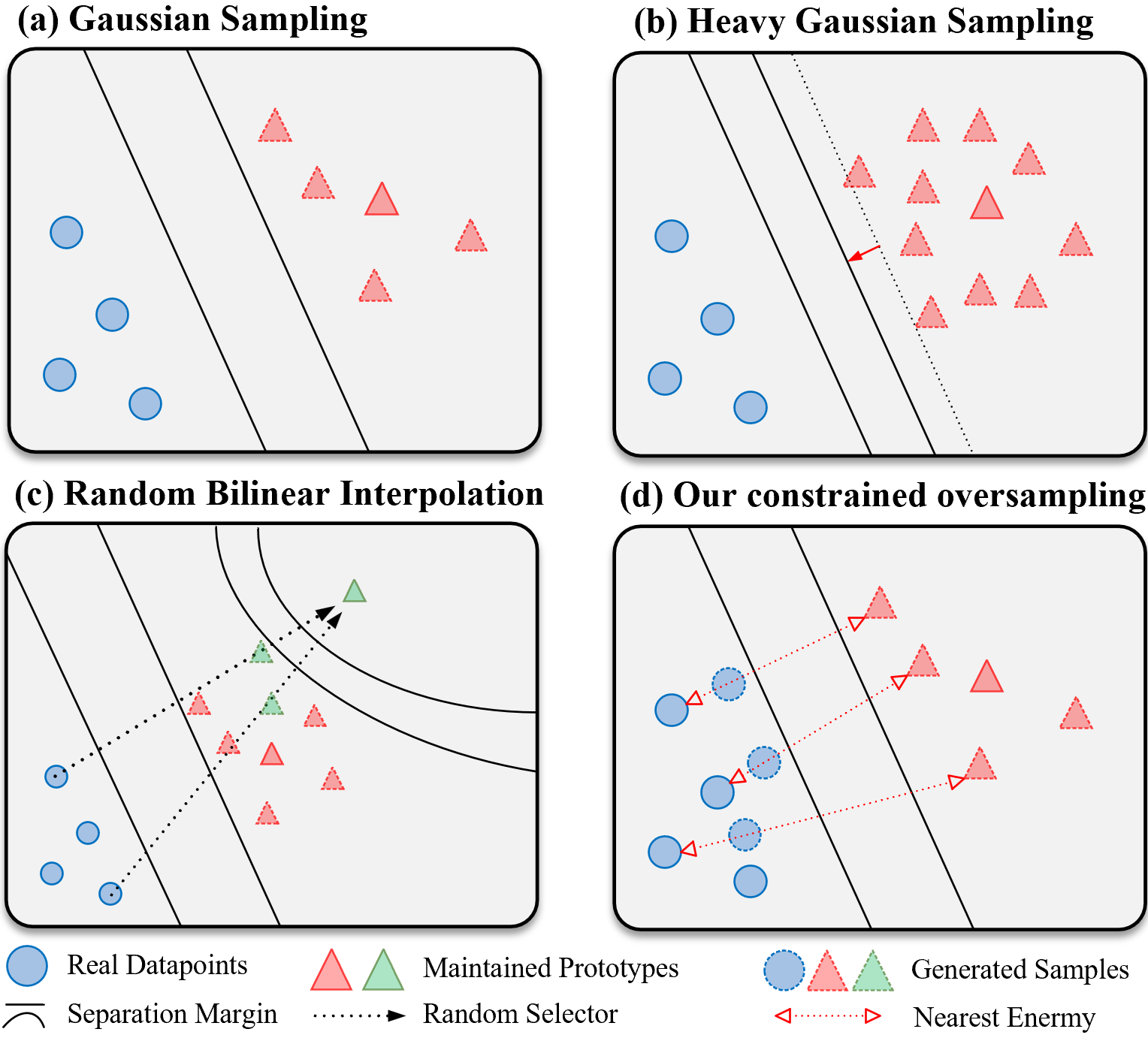}
  \vspace{-6mm}
  \caption{Naïve prototype rehearsal fails: (a–b) Gaussian sampling \cite{zhu2021pass,Magistri2024EFC} over-concentrates near old prototypes (prior inflation, poor boundary coverage); (c) random prototype interpolation overlaps class manifolds (PRAKA~\cite{shi2023prototype}); (d) our constrained oversampling moves uses information about nearest enemy, preserves prototype dominance and class separation margin \cite{NgnaweSPP:2024}.}
  \vspace{-7mm}
  \label{fig:first}
\end{figure} 

\begin{figure*}[t]  
  
    \centering
    \includegraphics[width=1\linewidth]{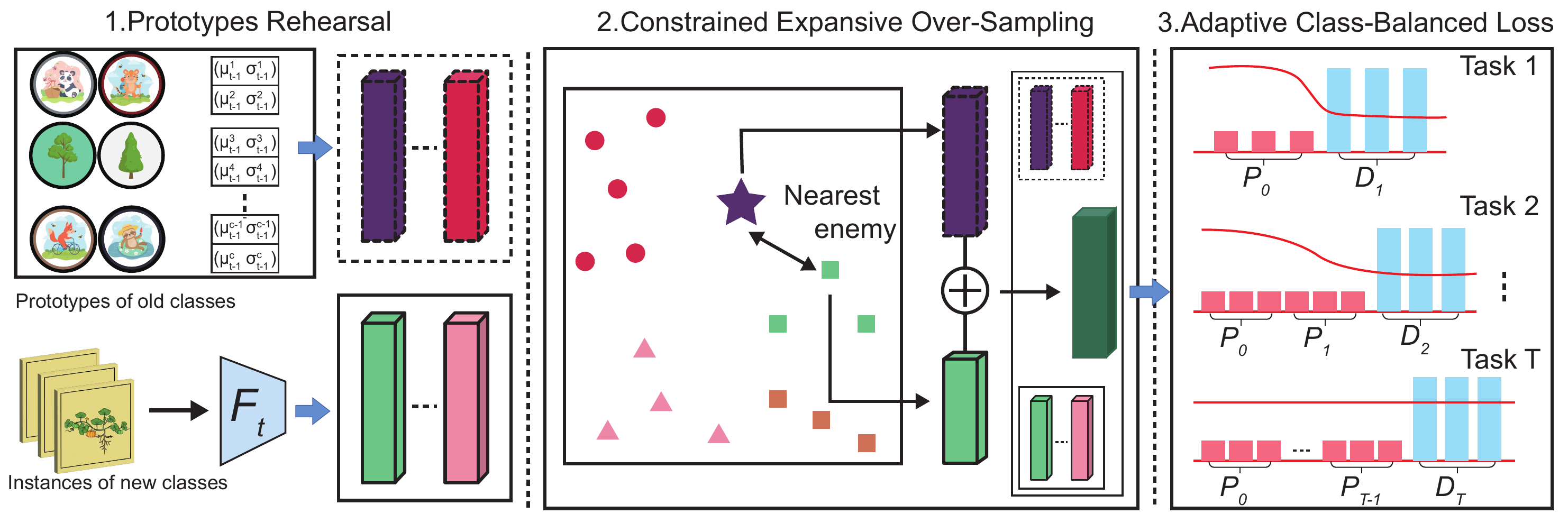}
    
    \caption{\textbf{Illustration of pipeline at task \(t\).} 
        \textbf{(1) Prototypes Rehearsal.}  
        One Gaussian
        \(\mathcal{N}(\mu_{\,t-1},\sigma_{\,t-1})\) prototype is maintained for every old class via drift compensation.
        At task \(t\) we randomly sample multiple features from the collection of prototypes and concatenate them with the
        current-task embeddings extracted by the backbone \(F_t\).
        \textbf{(2) Constrained Expansive Over-Sampling (CEOS).}  
        For every prototype feature we find its \(k\) nearest
        enemy features and interpolate them to generate synthetic minority examples that expand support up to—but not beyond—the decision boundary.
        \textbf{(3) Adaptive Class-Balanced (ACB) Loss.}  
        A time-dependent weight schedule (red curve) initially amplifies
        prototype gradients in early training stages, then gradually flattens; by the final task \(T\) the weights are nearly uniform across all classes, ensuring a balanced contribution from both replayed prototypes and current data.}
    \label{fig:main_figure}
\vspace{-5mm}
\end{figure*}

Continual learning (CL) \cite{WangZSZ:2024} aims to train models on task streams without retraining from scratch or catastrophically forgetting past knowledge \cite{WangYSH:2025}. A common baseline is memory-based rehearsal (storing and replaying past examples) \cite{Krutsylo:2024}. However, this conflicts with privacy constraints \cite{CaselliMBBAB:2024} and tight memory budgets \cite{ZhouTLSLYZ:2025} in many deployments. This motivates exemplar-free class-incremental learning (EFCIL) \cite{RoyVVG0D:2023,LiuZBLC:2024,WangG:2025}, where the learner must extend its label space without retaining raw images, relying instead on compact summaries such as class prototypes \cite{li2024steering,LiZHCW:2025,PhamH0DDN:2025,Sun0X00QL0:2025}.
Within EFCIL, two main families of methods have emerged. Prototype rehearsal \cite{zhu2021pass,shi2023prototype,Magistri2024EFC} stores a few representative prototypes per class and later samples synthetic features around them to approximate replay. Drift-compensation methods \cite{0004TLHWCJ020,GomezVillaGWBTW24,GoswamiSLKT024} instead realign stale prototypes with the evolving feature space, typically under a nearest-class-mean or softmax classifier. Recent benchmarks show drift compensation consistently outperforming prototype rehearsal \cite{rypesc2024task,GoswamiSLKT024}, reinforcing the view that prototype-based replay is inherently weaker.

We argue that this gap stems not from the principle of prototype rehearsal, but from how it is instantiated (see Figure~\ref{fig:first}). Existing pipelines have two key blind spots. First, prototypes are treated as isolated summaries: synthetic samples are drawn independently around each prototype, ignoring the geometry of nearby classes--specially nearest enemies that shape decision boundaries \cite{Gonzalez:2017,dablain2023efficient}. Second, rehearsal operates under a growing imbalance \cite{Ghosh:2024}: each old class is represented by a few synthetic points near a single prototype, while new classes contribute many real embeddings per task. Even in globally balanced datasets, this mismatch in effective sample counts biases the classifier toward recent tasks and erodes performance on early ones.

Standard Gaussian sampling exacerbates these issues \cite{Krawczyk:2020}. As the encoder adapts to new tasks, the embedding space drifts; sampling from a fixed spherical distribution around a stale prototype increasingly produces off-manifold features that no longer follow the true class manifold or its margins. Prototype rehearsal is thus penalized twice: it lacks informative cross-class geometry and operates in a temporally expanding long-tail regime where new classes dominate gradient updates. Drift compensation improves prototype alignment but does not address the missing enemy-aware information or the underlying imbalance.

We therefore revisit prototype rehearsal from a manifold- and imbalance-aware perspective. Our hypothesis is that prototype rehearsal can again be competitive---sometimes superior to drift-compensation-only pipelines---if it is redesigned to (i) explicitly exploit nearest-enemy information when generating rehearsal samples and (ii) correct the evolving imbalance between old prototypes and new classes through principled, time-based loss reweighting.

To this end, we introduce an exemplar-free EFCIL framework that couples manifold-aware prototype rehearsal with a temporal class-balancing loss (see Figure~\ref{fig:main_figure}) that closes—and often reverses—the gap to recent drift-compensation methods:
\begin{itemize}
    \item \textbf{Constrained Expansive Over-Sampling (CEOS).} We propose a boundary-aware oversampling scheme that augments each old-class prototype using its nearest enemy features from current tasks. Instead of fixed Gaussian noise, CEOS interpolates prototypes toward selected enemies along the data manifold, while enforcing margin-preserving constraints that keep synthetic points on the correct side of the decision boundary. This yields manifold-consistent, boundary-focused rehearsal samples that expand old-class support without collapsing into neighbors.
    \item \textbf{Adaptive Class-Balanced (ACB) Loss.} To counteract the hidden imbalance between few synthetic old-class features and many real new-class examples, we design a time-aware loss with temporal class weighting. Each class is assigned a virtual sample count that grows with its age, and its loss contribution is inversely scaled by this quantity, boosting gradients from freshly created prototypes and gradually attenuating them as richer supervision accumulates.
\end{itemize}

\section{Preliminaries}

\subsection{Problem Definition}
\label{sec:problem_def}

Continual learning (CL) studies how to update a model over a sequence of tasks while retaining prior knowledge.
In the \textbf{class-incremental} regime considered here, each task
$t \in \{1,\dots,T\}$ contributes a disjoint label set $\mathcal{C}_t$
with $\mathcal{C}_i \cap \mathcal{C}_j = \varnothing$ for $i \neq j$.
After finishing task~$t$, the learner must recognize any label in
$\mathcal{C}_{1:t} := \bigcup_{i=1}^{t} \mathcal{C}_i$
\emph{without} a task identifier at inference.

Let $F_t : \mathcal{X} \to \mathbb{R}^d$ denote the feature extractor after completing the first $t$ tasks.
During training on task~$t$, supervision is available only for
$D_t = \{(x_i, y_i) \mid y_i \in \mathcal{C}_t\}$.
The lack of access to past-task data defines the exemplar-free class-incremental setting.

\subsection{Prototype-Based Exemplar-Free CIL}
\label{sec:efcil}

In \textbf{exemplar-free class-incremental learning (EFCIL)}, raw inputs
from earlier tasks are not retained.  Focusing on a single transition
\(t{-}1 \!\to\! t\), the learner summarizes its history at the end of
task~\(t{-}1\) with per-class prototypes, i.e., empirical means for
each \(c\in\mathcal{C}_{1:t-1}\):
\begin{equation}
\label{eq:proto_def}
\bm{\mu}_c^{\,t-1}
=\frac{1}{|\mathcal{D}_c|}
\sum_{x\in\mathcal{D}_c} F_{t-1}(x),
\end{equation}
where \(\mathcal{D}_c\) collects all observed examples of class~\(c\)
up to step~\(t{-}1\).  During task \(t\), these cached statistics are the
only vestige of previous data, with a compact memory cost of
\(\mathcal{O}(|\mathcal{C}_{1:t-1}|\,d)\) floats for feature dimension \(d\).

Built on such summaries, several recent methods
\cite{0004TLHWCJ020,GoswamiSLKT024,GomezVillaGWBTW24}
use the nearest-class-mean decision rule:
\vspace{-2mm}
\begin{equation}
\hat{y}(x)=
\argmin_{c\in\mathcal{C}_{1:t-1}}
\bigl\|F_{t-1}(x)-\bm{\mu}_c^{\,t-1}\bigr\|_2^{2},
\end{equation}
incurring no additional trainable parameters.

\noindent\textbf{Prototype drift.}
When training on task \(t\), the encoder is updated from
\(F_{t-1}\) to \(F_{t}\) using labels exclusively from
\(\mathcal{C}_{t}\).  This reallocation of representational capacity to
separate new classes inevitably reshapes the embedding geometry.  As a
result, each previously stored prototype \(\bm{\mu}_c^{\,t-1}\)—defined
under \(F_{t-1}\)—is evaluated in the altered coordinates induced by
\(F_t\).  Let
\vspace{-1mm}
\begin{equation}
\bm{\mu}_c^{\,t} \;=\;
\frac{1}{|\mathcal{D}_c|}
\sum_{x\in\mathcal{D}_c} F_{t}(x),
\quad
\delta_c^{\,t} \;=\;
\bigl\|\bm{\mu}_c^{\,t}-\bm{\mu}_c^{\,t-1}\bigr\|_2
\end{equation}
denote the true class mean under the updated encoder and its displacement
from the stale prototype.  The scalar \(\delta_c^{\,t}\) quantifies
prototype drift: larger values indicate stronger misalignment. Drift yields two adverse effects:
(i) nearest-prototype predictions become biased toward the most recently
trained classes, degrading performance on earlier tasks;  
(ii) because old raw samples are unavailable, the learner cannot
recompute \(\bm{\mu}_c^{\,t}\) directly.  Hence, mitigating
\(\delta_c^{\,t}\)—or compensating for it—without violating the
exemplar-free constraint is a central objective of this work.

\subsection{Prototype Rehearsal}
\label{sec:proto_objectives}

Prototype rehearsal augments the current-task training data with
synthetic features generated from stored class prototypes. Given a
mini-batch of current-task samples
$B_t=\{(x_i,y_i)\}$ and a set of (optionally perturbed) prototypes
$\{\tilde{\bm{\mu}}_c^{\,t-1}\}$ for previously seen classes, a common formulation uses a cross-entropy loss that jointly trains on real and prototype-based features:
\vspace{-1mm}
\begin{equation}
\mathcal{L}_{\text{rehearsal}}
=
\underbrace{\mathcal{L}_{\text{CE}}\!\bigl(B_t,\;\mathcal{C}_t\bigr)}_{\text{fit new classes}}
+
\underbrace{\mathcal{L}_{\text{CE}}\!\bigl(B_t\cup\{\tilde{\bm{\mu}}_c^{\,t-1}\},\;
               \mathcal{C}_{\le t}\bigr)}_{\text{rehearse all seen classes}},
\label{eq:prace}
\end{equation}
where the first term adapts the model to the newly introduced classes
$\mathcal{C}_t$, and the second term uses both current samples and
prototype features to reinforce the decision boundaries over all classes
observed so far, $\mathcal{C}_{\le t}$.

\section{Limitations of current prototype rehearsal strategies.}
\label{sec:limitations}
The way prototype rehearsal is typically instantiated introduces structural weaknesses that explain its performance gap to recent drift-compensation methods.

\noindent \textbf{Limitations of Gaussian-based sampling.} Most approaches treat prototypes as isolated class summaries and sample synthetic features around each prototype independently, usually from a spherical Gaussian. This causes prior inflation: samples concentrate in a narrow cluster around the prototype and fail to populate the regions where decision boundaries are formed. As shown in Figure~\ref{fig:first}(a–b), such sampling offers poor boundary coverage and ignores the local geometry of enemy classes, so rehearsal mostly reinforces “easy’’ regions \cite{DablainBKAC:2024} rather than contested margins.

\noindent \textbf{Impact on class separability.} Boundary-agnostic interpolation can also harm separability. Methods that interpolate between prototypes or between prototypes and new features without explicit enemy awareness often generate mixed embeddings that cross class manifolds. When the interpolation direction is not aligned with the nearest enemy, synthetic points can overlap with neighboring classes, collapsing margins instead of sharpening them, as illustrated by the random bi-interpolation behavior in Figure~\ref{fig:first}(c). As the feature space drifts over tasks, Gaussian or random interpolation becomes increasingly misaligned with the true class manifolds, and rehearsal samples tend to be off-manifold.

\noindent \textbf{Hidden imbalance in prototype rehearsal.} Current pipelines are blind to the evolving imbalance between old and new classes. Each old class is represented by a few synthetic features near a single prototype, while new classes contribute many real embeddings at every task. Even under globally balanced streams, this induces a long-tail regime in which early classes form a thin, noisy tail and recent classes dominate gradient updates. Prototype rehearsal thus operates where (i) old classes are under-represented in both quantity and geometric coverage, and (ii) their prototypes gradually drift away from true class means, yielding unreliable rehearsal samples. These factors bias the classifier toward recent tasks and cause early-class performance to erode, even when drift compensation is present.

\noindent \textbf{Need for revisiting prototype rehearsal.} Overall, standard prototype rehearsal induces a structural bias: old classes rely on few, narrowing, drifting synthetic samples, whereas new classes receive dense, up-to-date supervision. Even for globally balanced streams, this mismatch creates a temporal long tail and misaligned prototypes. In the next section, we formalize this prototype-induced imbalance and show how it systematically biases the softmax classifier toward recent classes and degrades rehearsal under semantic drift.


\section{Uncovering Prototype–Induced Imbalance}
\label{sec:imbalance}
Building on the failure modes identified in Section~\ref{sec:limitations}, we now formalize how prototype rehearsal induces a temporal class imbalance, even when each task contributes a perfectly balanced stream. The compression of old data into a few prototype-centered synthetic samples and the accumulation of feature drift combine to yield an increasingly skewed and misaligned training signal for the classifier.

Imbalance is a significant challenge in machine learning
\cite{YangJSG:2022,Zhang:2023,DablainBKAC:2024,Ghosh:2024}.  Prior CL studies discuss it mainly as skew across or
within tasks \cite{LiuHCBLC:2022,He:2024,Wang:2024,GuYYWZGD:2025,Zhang:2025}, including exemplar-free settings
\cite{Wang:2024}.  The imbalance we identify in prototype rehearsal for EFCIL is different: it
\emph{accumulates over time} even when the incoming stream is balanced
per task.  In this sense, it resembles data-stream imbalance
\cite{Aguiar:2024}, where evolving distributions and online updates
compound long-run effects.

\noindent \textbf{Skewed training of the classification head.}
The resulting per-class skew mirrors long-tail recognition
\cite{Zhang:2023}: majority classes dominate gradients and tilt decision
boundaries \cite{Ghosh:2024}.  Here, abundant new-class samples and
sparse rehearsal for old classes yield much richer gradients for recent
labels, biasing updates toward new knowledge and progressively diluting
the signal from earlier classes. 

\noindent \textbf{Theorem 1 (Asymptotic Softmax Bias during Prototype Rehearsal)}.
Let $\{ \mathcal{T}_t \}_{t=1}^T$ be a sequence of $T$ supervised
classification tasks. Each task $\mathcal{T}_t$ introduces $C$ new
classes with $K > 1$ i.i.d.\ samples per class drawn from
$\{ \mathcal{D}_c \}_{c \in \mathcal{C}_t}$. Let
$f: \mathcal{X} \to \mathbb{R}^d$ be a fixed unit-norm embedding, and
let $W_t \in \mathbb{R}^{C_{\leq t} \times d}$ be the weight matrix of a
linear classifier trained with the softmax cross-entropy:
\vspace{-1mm}
\begin{equation}
\scalebox{0.92}{$%
  \displaystyle
  \mathcal{L}_t(W_t) = -\frac{1}{N_t}
  \sum_{(x,y)\in \mathcal{S}_t \cup \mathcal{M}_{t-1}}
  \log
  \left(
    \frac{\exp(W_{t,y}^{\top} f(x))}
         {\sum_{j\in\mathcal{C}_{\le t}}
          \exp(W_{t,j}^{\top} f(x)) }
  \right)
$}
\label{eq:ce_loss}
\end{equation}
where $\mathcal{S}_t$ is the dataset of task $t$, and $\mathcal{M}_{t-1}$
is a fixed-size memory containing exactly $m \geq 1$ i.i.d.\ exemplars
per old class drawn from each $\mathcal{D}_c$.

Assume: (i) $f(x)$ is unit-norm for all $x$; (ii) full-batch gradient
descent with a fixed number of epochs per task; (iii) sampling from a
single prototype so that every old class has exactly $m$ replay
instances; (iv) constant $C$ and $K$ per task; and (v) $m \ll K$
(extreme imbalance). Then for any $c \in \mathcal{C}_{\leq T-1}$,
\vspace{-1mm}
\begin{equation}
\label{eq:vanish}
\lim_{T \to \infty} \mathbb{E}_{x \sim \mathcal{D}_c} \left[ p_{W_T}(y = c \mid x) \right] = 0.
\end{equation}
\vspace{-3mm}

\noindent \textbf{Degrading quality of rehearsal sampling.}
Prototype rehearsal typically draws synthetic embeddings for an old class
from a Gaussian centered at its stored prototype (mean or distribution).
This parallels oversampling under extreme imbalance, where minority
classes are augmented with artificial instances.  A considerable body of
work argues that Gaussian sampling poorly respects class manifolds
\cite{Bellinger:2018,dablain2023efficient}; nevertheless, Gaussian
assumptions are common in prototype rehearsal \cite{Magistri2024EFC}.
We contend that this choice degrades over time as drift increases.

\noindent \textbf{Theorem 2 (Degradation of Gaussian Sampling for Prototype Rehearsal under Drift)}.
Assume: (i) for $t' > t$, rehearsal for class $c$ samples
$\tilde{x} \sim \mathcal{N}(\mu_c^{(t)}, \Sigma_c^{(t)})$; (ii) due to
embedding drift, the true mean evolves so that
$\|\mu_c^{(t')} - \mu_c^{(t)}\|_2 \geq \delta_t$ with
$\delta_t \to \infty$ as $t \to \infty$. Then the expected logit
alignment between $W_{T,c}$ and synthetic samples decays:
\vspace{-1mm}
\begin{equation}
\lim_{T \to \infty} \mathbb{E}_{\tilde{x} \sim \mathcal{N}(\mu_c^{(t)}, \Sigma_c^{(t)})} \left[ W_{T,c}^\top \tilde{x} \right] = 0.
\end{equation}
\vspace{-2mm}

Combined with Eq.~\ref{eq:vanish}, this implies the classifier’s
posterior for old class $c$ on real data vanishes asymptotically.  Full
proofs for both theorems are provided in the Appendix.

In summary, absent explicit balancing mechanisms, prototype-rehearsal
EFCIL models tend to overfit to newly introduced tasks and under-represent
earlier ones—even when catastrophic forgetting and prototype drift are
partially controlled.

\section{Methodology}\label{Method}

To neutralize the issues with prototype rehearsal discussed in
preceding sections, we propose a new rehearsal strategy that
remains strictly within the exemplar-free budget (see Figure~\ref{fig:main_figure}).  
\textbf{First}, a \textbf{Constrained Expansive Over-Sampling (CEOS)} routine
augments minority support in embedding space by interpolating prototypes
toward carefully chosen enemy features, broadening class coverage and focusing on difficult instances \cite{DablainBKAC:2024}
without crossing decision boundaries.  
\textbf{Second}, an \textbf{Adaptive Class-Balanced (ACB) Loss} dynamically
re-weights each class so that freshly minted prototypes exert strong
influence only when they are most representative, then gradually cede
importance to the richer manifold of incoming data.  
The following subsections detail these two complementary modules.

\subsection{Constrained Expansive Over-Sampling}

\label{sec:ceos}
Inspired by embedding-space augmentation techniques
\cite{verma2019manifold,zhong2021mislas,park2022majority,dablain2023efficient},
we synthesize minority features by interpolating each minority prototype
$\mathbf{p}\!\in\!\mathbb{R}^d$ with one of its nearest enemy
neighbors $\mathbf{e}\!\in\!\mathbb{R}^d$ (a different-class sample
from the same mini-batch).  
For a prototype–enemy pair $(\mathbf{p},\mathbf{e})$, we draw a mixing
coefficient
\vspace{-1mm}
\begin{equation}
  \lambda\;\sim\;\mathcal{D},
  \qquad
  \text{with}\;\; \lambda\in(\tau,1),\; \tau>\tfrac12,
\end{equation}
\vspace{-1mm}
where $\mathcal{D}$ is any continuous distribution supported on
$(\tau,1)$; the lower bound $\tau$ guarantees prototype dominance.
The synthetic feature is the convex combination
\vspace{-1mm}
\begin{equation}
  \tilde{\mathbf{x}}
  \;=\;
  \lambda\,\mathbf{p} \;+\; (1-\lambda)\,\mathbf{e},
  \label{eq:ceos_mix}
\end{equation}
\vspace{-1mm}
and is assigned the hard label of $\mathbf{p}$.

The augmentation routine proceeds in three steps:
\textbf{(i)} for every minority prototype, retrieve up to $k$ nearest
enemies by Euclidean distance in the current feature space;
\textbf{(ii)} for each prototype–enemy pair, sample an independent
$\lambda$ and apply~\eqref{eq:ceos_mix}; and
\textbf{(iii)} add the resulting synthetic features to the mini-batch.
Because $\lambda>\tfrac12$, every synthetic point lies on the interior
segment connecting $\mathbf{p}$ to $\mathbf{e}$ and remains closer to
$\mathbf{p}$ than to $\mathbf{e}$, thereby expanding minority
support toward the decision boundary without crossing it.

\noindent \textbf{Estimating mixing coefficient.} We define a local decision boundary between the prototype \textbf{p} and enemy \textbf{e} using a Mahalanobis distance $d_M(\mathbf{x}, \mathbf{p}) = \sqrt{(\mathbf{x} - \mathbf{p})^\top \Sigma^{-1} (\mathbf{x} - \mathbf{p})}$, where $\Sigma$ is estimated from the current mini-batch of instances. We are looking for such a value of $\lambda$ that the synthetic point $\tilde{\mathbf{x}}$ remains on the correct side of the margin, i.e., $d_M(\tilde{\mathbf{x}}, \mathbf{p}) < d_M(\tilde{\mathbf{x}}, \mathbf{e})$. This leads to a lower bound \(\tau\) on \(\lambda\) that can be computed analytically:
\begin{equation}
    \lambda > \frac{1}{2} + \frac{\Delta^\top \Sigma^{-1} \delta}{2\, \delta^\top \Sigma^{-1} \delta}, \quad \text{where} \quad \Delta = \mathbf{e} - \mathbf{p}, \; \delta = \tilde{\mathbf{x}} - \mathbf{p}
\end{equation}

Solving this leads to a closed-form $\tau$ that guarantees the synthetic point lies within the prototype-dominant space. This allows for a two-fold effective oversampling: (i) \textbf{Mixing coefficient selection:} sampling $\lambda$ from the interval $(\tau, 1)$ with $\tau$ defined as above ensures both class consistency and margin preservation; and (ii) \textbf{Adaptive mixing coefficient:} for each \((\mathbf{p}, \mathbf{e})\), compute a local safe lower bound $\tau_{(\mathbf{p}, \mathbf{e})}$ using margin analysis, and sample $\lambda \sim \mathcal{U}(\tau_{(\mathbf{p}, \mathbf{e})}, 1)$. Full derivation of the coefficient is shown in the Appendix.


\subsection{Adaptive Class-Balanced Loss}

\label{sec:acb}
We now introduce an \textbf{Adaptive Class-Balanced Loss} that regulates
the relative influence of old and new classes during
class-incremental training.
Each mini-batch mixes two sources of supervision: (i) real features
from the current task and (ii) prototype features sampled from
the per-class Gaussian accumulated in previous tasks.
Although both streams have equal batch size, prototypes form the
minority signal because each is sampled from a \textbf{single,
uni-modal distribution} representing only the class mean and local
variance, whereas real samples cover the richer manifold of the incoming
data.
If these narrowly distributed prototypes were always given a large,
static weight, the optimizer would over-attend to them and under-fit the
current data.
To prevent this imbalance we let every old class accumulate a larger
virtual sample count over time, thereby reducing its
effective weight in the loss.

For class $c$ first encountered at task $t_c$, its virtual sample count
at task $t\!\ge\!t_c$ is
\begin{equation}
N_c(t)=\min\Bigl\{N_{\max},
\,N_{\min}+\bigl(N_{\max}-N_{\min}\bigr)
      \bigl(\tfrac{t-t_c}{T}\bigr)^{\gamma}\Bigr\},
\label{eq:asc_samples}
\end{equation}
where $N_{\min}$ and $N_{\max}$ are the start and cap values, $T$
normalizes the horizon (e.g., the total number of tasks), and
$\gamma>0$ controls the growth speed.
The corresponding class-balanced weight
\vspace{-1mm}
\begin{equation}
w_c(t)=\frac{1-\beta}{1-\beta^{N_c(t)}},\qquad 0<\beta<1,
\label{eq:cb_weight}
\end{equation}
\vspace{-1mm}
is monotonically decreasing in $N_c(t)$; for $\beta\!\to\!1$ we
have $w_c(t)\!\approx\!1/N_c(t)$.
Thus, as a class ages, its virtual sample count ascends and its
loss weight descends, ensuring that prototypes are influential
immediately after creation—when they best approximate the evolving
decision boundary—but become progressively less dominant as newer, more
diverse data arrive.

We incorporate these weights into a cross-entropy objective
\vspace{-1mm}
\begin{equation}
\mathcal{L}_{\text{ACB}}(t)=
-\frac{1}{|\mathcal{B}|}\sum_{(x,y)\in\mathcal{B}}
            w_{y}(t)\,\log p_{y}(x),
\label{eq:acbl_loss}
\end{equation}
\vspace{-1mm}
where each mini-batch $\mathcal{B}$ contains both prototype and
current-task features.
By coupling the ascending virtual sample schedule with the inverse
class-balanced weighting, $\mathcal{L}_{\text{ACB}}$ automatically shifts
attention away from early, single-model prototypes toward the richer data
of new tasks, realizing a principled stability–plasticity trade-off
without retaining any raw exemplars.

\vspace{-1mm}
\section{Experiments}
\vspace{-1mm}
\label{sec:exp}

\begin{table*}[t]
\caption{Average incremental ($A_{\text{inc}}$) and last ($A_{\text{last}}$)
accuracy (\%, mean $\pm$ std.\ over five runs) on CIFAR-100 and TinyImageNet when training the feature
extractor from scratch. Best results are \textbf{bold}. $^\ddagger$ marks prototype rehearsal approaches.}
\small    
\setlength{\tabcolsep}{4.2pt}
\centering
\begin{tabular}{lcccccccccc}
\toprule
\multirow{3}{*}{\textbf{Method}} &
  \multicolumn{4}{c}{\textbf{CIFAR‑100}} &
  \multicolumn{6}{c}{\textbf{TinyImageNet}} \\
\cmidrule(lr){2-5}\cmidrule(lr){6-11}
  & \multicolumn{2}{c}{$T{=}10$} &
    \multicolumn{2}{c}{$T{=}20$} &
    \multicolumn{2}{c}{$T{=}10$} &
    \multicolumn{2}{c}{$T{=}20$} &
    \multicolumn{2}{c}{$T{=}40$} \\
\cmidrule(lr){2-3}\cmidrule(lr){4-5}
\cmidrule(lr){6-7}\cmidrule(lr){8-9}\cmidrule(lr){10-11}
  & $A_{\text{last}}$ & $A_{\text{inc}}$
  & $A_{\text{last}}$ & $A_{\text{inc}}$
  & $A_{\text{last}}$ & $A_{\text{inc}}$
  & $A_{\text{last}}$ & $A_{\text{inc}}$
  & $A_{\text{last}}$ & $A_{\text{inc}}$ \\
\midrule
EWC &
  30.9$\pm$1.9 & 50.4$\pm$1.7 &
  17.0$\pm$1.6 & 34.2$\pm$2.1 &
  18.5$\pm$1.8 & 34.3$\pm$2.3 &
  11.3$\pm$1.9 & 26.8$\pm$2.5 &
   3.1$\pm$1.5 & 12.3$\pm$2.1 \\[2pt]

LwF\conf{ECCV16} &
  31.9$\pm$1.1 & 51.8$\pm$1.5 &
  17.6$\pm$1.2 & 39.2$\pm$1.7 &
  27.1$\pm$1.5 & 39.6$\pm$2.0 &
  15.2$\pm$1.6 & 31.5$\pm$2.1 &
   6.0$\pm$1.2 & 18.3$\pm$1.6 \\[2pt]

SDC\conf{CVPR20} &
  40.6$\pm$0.9 & 56.2$\pm$1.3 &
  32.3$\pm$1.0 & 46.6$\pm$1.4 &
  29.5$\pm$1.1 & 43.8$\pm$1.5 &
  26.3$\pm$1.2 & 40.6$\pm$1.7 &
  10.9$\pm$1.0 & 24.8$\pm$1.4 \\[2pt]

PASS\conf{CVPR21}$^\ddagger$ &
  30.8$\pm$1.2 & 48.3$\pm$1.1 &
  17.6$\pm$0.8 & 31.1$\pm$1.3 &
  24.5$\pm$0.6 & 39.5$\pm$1.0 &
  18.5$\pm$1.4 & 30.4$\pm$1.9 &
  7.2$\pm$2.2 & 20.4$\pm$2.5 \\[2pt]

FeTrIL\conf{WACV23}$^\ddagger$ &
  34.9$\pm$0.5 & 51.2$\pm$1.1 &
  23.3$\pm$1.4 & 37.9$\pm$1.2 &
  31.0$\pm$0.9 & 45.3$\pm$1.8 &
  25.9$\pm$1.2 & 39.9$\pm$1.2 &
  10.3$\pm$1.6 & 22.9$\pm$1.7 \\[2pt]

PRAKA\conf{ICCV23}$^\ddagger$ &
  42.4$\pm$1.3 & 56.9$\pm$1.6 &
  31.3$\pm$1.1 & 46.2$\pm$1.5 &
  29.7$\pm$1.6 & 40.7$\pm$1.4 &
  24.4$\pm$1.5 & 35.6$\pm$1.0 &
  9.4$\pm$1.0 & 22.5$\pm$0.7 \\[2pt]

FeCAM\conf{NIPS23} &
  32.4$\pm$0.5 & 48.7$\pm$0.9 &
  21.1$\pm$1.0 & 34.5$\pm$1.3 &
  30.9$\pm$0.9 & 44.9$\pm$1.4 &
  24.9$\pm$0.8 & 37.9$\pm$1.4 &
  11.4$\pm$0.5 & 24.3$\pm$1.1 \\[2pt]

EFC\conf{ICLR24}$^\ddagger$ &
  43.5$\pm$0.8 & 58.1$\pm$1.2 &
  32.4$\pm$0.9 & 47.0$\pm$1.3 &
  34.5$\pm$1.1 & 47.9$\pm$1.5 &
  28.4$\pm$1.2 & 42.1$\pm$1.6 &
  21.5$\pm$1.1 & 34.8$\pm$1.5 \\[2pt]

ADC\conf{CVPR24} &
  45.2$\pm$1.2 & 59.7$\pm$1.6 &
  35.1$\pm$1.4 & 51.7$\pm$1.8 &
  30.2$\pm$1.5 & 42.4$\pm$1.9 &
  18.1$\pm$1.6 & 36.0$\pm$2.1 &
  13.6$\pm$1.4 & 26.8$\pm$1.8 \\[2pt]

LDC\conf{ECCV24} &
  45.4$\pm$1.6 & 59.5$\pm$1.9 &
  \textbf{35.5$\pm$1.9} & \textbf{51.9$\pm$2.3} &
  34.2$\pm$1.1 & 46.8$\pm$1.6 &
  24.9$\pm$2.2 & 38.2$\pm$2.7 &
  15.3$\pm$1.7 & 29.7$\pm$1.9 \\[2pt]
\rowcolor{mygray!20}
Ours &
  \textbf{46.9$\pm$1.0} & \textbf{60.2$\pm$1.4} &
  34.9$\pm$1.6 & 48.0$\pm$1.3 &
  \textbf{35.8$\pm$0.9} & \textbf{49.0$\pm$1.4} &
  \textbf{31.8$\pm$1.0} & \textbf{44.3$\pm$1.4} &
  \textbf{23.2$\pm$0.8} & \textbf{36.3$\pm$1.2} \\
\bottomrule
\end{tabular}

\label{tab:main_results}
\end{table*}

\begin{table*}[t]
\caption{Quantitative results on ImageNet-100 and CUB-200. Best results are \textbf{bold}.\dag: results excerpted from~\cite{GomezVillaGWBTW24}. $^\ddagger$ marks prototype rehearsal approaches.}
\centering
\begin{tabular}{lcccccccc}
\toprule
\multirow{3}{*}{\textbf{Method}} &
  \multicolumn{4}{c}{\textbf{ImageNet-100}} &
  \multicolumn{4}{c}{\textbf{CUB-200}} \\
\cmidrule(lr){2-5}\cmidrule(lr){6-9}
  & \multicolumn{2}{c}{$T{=}10$} &
    \multicolumn{2}{c}{$T{=}20$} &
    \multicolumn{2}{c}{$T{=}10$} &
    \multicolumn{2}{c}{$T{=}20$} \\
\cmidrule(lr){2-3}\cmidrule(lr){4-5}
\cmidrule(lr){6-7}\cmidrule(lr){8-9}
  & $A_{\text{last}}$ & $A_{\text{inc}}$
  & $A_{\text{last}}$ & $A_{\text{inc}}$
  & $A_{\text{last}}$ & $A_{\text{inc}}$
  & $A_{\text{last}}$ & $A_{\text{inc}}$ \\
\midrule
EWC &
  25.1$\pm$2.8 & 40.6$\pm$3.3 &
  13.7$\pm$2.1 & 29.2$\pm$2.5 &
  15.8$\pm$0.7 & 32.6$\pm$0.5 &
  12.3$\pm$0.8 & 27.2$\pm$0.6 \\[2pt]

LwF\conf{ECCV16} &
  33.4$\pm$2.2 & 51.5$\pm$1.6 &
  18.6$\pm$1.6 & 41.3$\pm$1.9 &
  30.4$\pm$1.1 & 46.1$\pm$1.0 &
  19.4$\pm$1.6 & 34.7$\pm$1.8 \\[2pt]

SDC\conf{CVPR20} &
  35.4$\pm$1.9 & 50.1$\pm$1.6 &
  19.4$\pm$1.0 & 36.5$\pm$1.4 &
  50.3$\pm$1.3 & 60.5$\pm$1.2 &
  27.9$\pm$1.4 & 40.1$\pm$1.6 \\[2pt]

PASS\conf{CVPR21}$^\ddagger$ &
  26.4$\pm$1.3 & 45.7$\pm$0.2 &
  14.4$\pm$1.2 & 31.7$\pm$0.4 &
  27.0$\pm$0.9 & 42.3$\pm$0.9 &
  18.1$\pm$1.2 & 36.9$\pm$1.1 \\[2pt]

FeTrIL\conf{WACV21}$^\ddagger$ &
  36.2$\pm$1.2 & 52.6$\pm$0.6 &
  26.6$\pm$1.5 & 42.4$\pm$2.1 &
  36.9$\pm$0.7 & 48.2$\pm$0.6 &
  34.6$\pm$1.0 & 45.3$\pm$0.9 \\[2pt]

FeCAM\conf{NIPS23} &
  38.7$\pm$1.0 & 54.8$\pm$0.5 &
  29.0$\pm$1.3 & 44.6$\pm$2.0 &
  40.2$\pm$0.8 & 54.9$\pm$1.0 &
  36.2$\pm$1.1 & 48.9$\pm$1.3 \\[2pt]

EFC\conf{ICLR24}$^\ddagger$ &
  50.9$\pm$1.1 & 61.3$\pm$1.2 &
  38.6$\pm$1.2 & 50.5$\pm$1.5 &
  51.0$\pm$0.6 & 63.3$\pm$0.7 &
  46.1$\pm$1.0 & 59.3$\pm$1.3 \\[2pt]

ADC\conf{CVPR24} &
  38.3$\pm$1.2 & 55.5$\pm$1.5 &
  25.1$\pm$1.3 & 43.4$\pm$1.7 &
  49.5$\pm$0.9 & 58.8$\pm$1.1 &
  35.4$\pm$1.4 & 48.3$\pm$1.4 \\[2pt]

LDC\conf{ECCV24} &
  51.4\textsuperscript{\dag}$\pm$1.2\textsuperscript{\dag} & \textbf{69.4\textsuperscript{\dag}$\pm$0.6\textsuperscript{\dag}} &
  28.5$\pm$1.7 & 46.5$\pm$2.7 &
  47.5$\pm$0.7 & 55.7$\pm$1.3 &
  27.2$\pm$1.1 & 39.8$\pm$2.1 \\[2pt]
\rowcolor{mygray!20}
Ours &
  \textbf{52.7$\pm$1.4} & 65.1$\pm$1.1 &
  \textbf{41.1$\pm$1.3} & \textbf{53.0$\pm$1.9} &
  \textbf{54.1$\pm$0.8} & \textbf{65.7$\pm$1.1} &
  \textbf{48.1$\pm$1.0} & \textbf{61.0$\pm$1.3} \\
\bottomrule
\end{tabular}

\vspace{-6mm}
\label{tab:main_results2}
\end{table*}

\textbf{Baselines and hyper-parameters.}
We evaluate our method against multiple exemplar-free class-incremental learning (EFCIL) baselines.  
Classic methods—EWC~\cite{kirkpatrick2017ewc} and LwF~\cite{LiH16}—are run with the reference implementation from the \textsc{OCL} framework~\cite{MAI202228}.  
Recent state-of-the-art methods—SDC~\cite{0004TLHWCJ020}, PASS~\cite{zhu2021pass}, FeTrIL~\cite{petit2023fetril}, PRAKA~\cite{shi2023prototype}, FeCAM~\cite{toldo2022fecam}, EFC~\cite{Magistri2024EFC}, ADC~\cite{GoswamiSLKT024}, and LDC~\cite{GomezVillaGWBTW24}—are executed via the official code released in FACIL~\cite{masana2022class}, PyCIL~\cite{zhou2023pycil}, or the authors’ repositories.  
Unless stated otherwise, we retain each paper’s original data augmentations and default hyperparameters.

\noindent \textbf{Implementation details and reproducibility.}
Our codebase extends the public EFC implementation with the components introduced in this work.  
All experiments use a ResNet-18 backbone trained from scratch~\cite{he2016deep}.  
Following EFC, we adopt a batch of 64 real images and a companion batch of 64 prototypes.  
Both \textsc{CIFAR-100}~\cite{krizhevsky2009learning} and \textsc{TinyImageNet}~\cite{le2015tiny} are trained for 100 epochs with Adam, a fixed learning rate of $1\times10^{-4}$, and weight decay $2\times10^{-4}$. While for the \textsc{ImageNet-100}~\cite{5206848} and \textsc{CUB-200}~\cite{wah2011caltech}, we use a learning rate of $1\times10^{-5}$ for the backbone and $1\times10^{-4}$ for the heads.
In Eq.~\eqref{eq:asc_samples} we set the virtual-count bounds to \(N_{\min}=100\) and \(N_{\max}=500\).
CEOS selects one enemy per prototype ($k{=}1$), thus injecting 64 synthetic samples from the second task onward. Source code, and configuration files will be released upon publication to ensure full reproducibility. 

\noindent \textbf{Evaluation Metrics.} We report the two most common metrics: the \textbf{last‑task average accuracy} $A_{\text{last}}$ and its running mean, the  
\textbf{average incremental accuracy} $A_{\text{inc}}$. 
More details of the datasets, parameter settings and evaluation metrics are in the Appendix.

\vspace{-1mm}
\subsection{Main Results}
\vspace{-1mm}
Table~\ref{tab:main_results} and Table~\ref{tab:main_results2} summarize the training-from-scratch results of our method against all competitors.
Experiments are conducted on balanced task partitions of four benchmarks—
\textsc{CIFAR--100}, \textsc{TinyImageNet}, \textsc{ImageNet-100} and \textsc{CUB-200}.
We report the mean $\pm$ standard deviation of both accuracy metrics over five independent runs.

On \textbf{CIFAR--100}, our method leads the 10‑task split with \textbf{46.9}\,\% \(A_{\text{last}}\) and \textbf{60.2}\,\% \(A_{\text{inc}}\),
edging LDC by 1.5 / 0.7 percentage points (pp).
With 20 tasks, we post 34.9\,\% / 48.0\,\%, only 0.6 / 3.9 pp behind LDC
yet still 2.5/1.0 pp ahead of EFC -- evidence of a balanced plasticity–stability trade‑off. The advantage widens on \textbf{TinyImageNet}. The margin over LDC grows to
1.6 / 2.2 pp (10 tasks) and 6.9 / 6.1 pp (20 tasks). Under the more demanding 40‑task stream we still record \textbf{23.2}\,\% \(A_{\text{last}}\) and \textbf{36.3}\,\% \(A_{\text{inc}}\), leading LDC by 7.9 / 6.6 pp. Training from scratch on \textbf{ImageNet--100} (Table~\ref{tab:main_results2}),
we achieve \textbf{52.7}\,\% / \textbf{65.1}\,\% on the 10‑task split,
surpassing EFC by 1.8 / 3.8 pp and the published LDC numbers by
1.3 pp in \(A_{\text{last}}\) (but 4.3 pp lower in \(A_{\text{inc}}\)).
\ul{Our rerun of public LDC code yields}
\(A_{\text{last}} = 41.7\pm1.5\,\%\) and
\(A_{\text{inc}} = 58.7\pm1.7\,\%\). On \textbf{CUB--200} with ImageNet pre‑training, our method improves on the
second‑best approach by 2.0 / 1.7 pp (20 tasks).

\begin{figure}[t]
  \centering
  \includegraphics[width=.85\columnwidth]{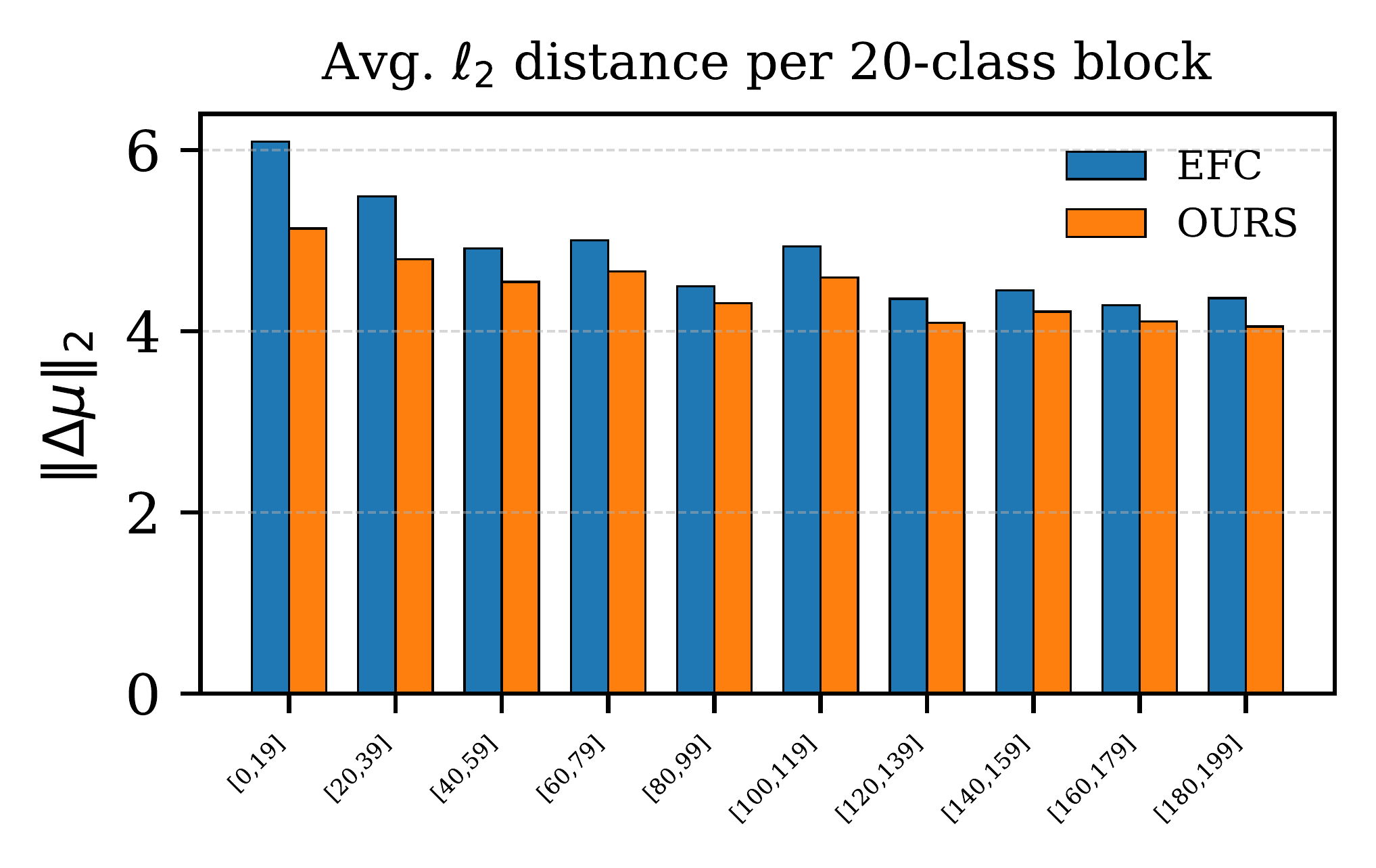}
  \vspace{-2mm}
  \caption{Tiny~ImageNet ($T{=}20$): The average distance between each class prototype and its real class mean.}
  \label{fig:drift}
  \vspace{-10pt} 
\end{figure}

Strictly speaking, our method is not designed to reduce prototype drift directly. Nevertheless, quantitative analysis of the models trained on Tiny‑ImageNet shows that (Figure~\ref{fig:drift}), in practice, it results in noticeably less drift of the prototypes overall.

These consistent gains show that our adaptive class-balanced loss, coupled with expansive oversampling,
mitigates distribution drift more effectively than the other SOTA methods.
In summary, our method simultaneously improves old-class retention and new-class acquisition,
achieving state-of-the-art accuracy across most splits and scaling robustly as the task sequence lengthens.

\setlength{\tabcolsep}{6pt}

\begin{table}[t]
  \caption{Tiny\,ImageNet ($T{=}10$ and $T{=}20$): Contributions of different components.}
  
  \centering
  \small
  \setlength{\tabcolsep}{4pt}
  \renewcommand{\arraystretch}{0.9}
  \resizebox{\columnwidth}{!}{%
    \begin{tabular}{@{}ccc|cc|cc@{}}
      \toprule
        \multicolumn{3}{c|}{\textbf{Components}} &
          \multicolumn{2}{c|}{$T{=}10$} &
          \multicolumn{2}{c}{$T{=}20$} \\
        CEOS & $L_{\text{acb}}$ & $L_{\text{focal}}$
            & $A_{\text{last}}$(\%) & $A_{\text{inc}}$(\%)
            & $A_{\text{last}}$(\%) & $A_{\text{inc}}$(\%) \\
      \midrule
        \xmark & \xmark & \xmark & 34.5$\pm$1.1 & 47.9$\pm$1.6 & 28.4$\pm$1.2 & 42.1$\pm$1.6\\
        \cmark & \xmark & \xmark & 35.1$\pm$1.2 & 48.7$\pm$1.5 & 30.4$\pm$1.1 & 43.0$\pm$1.4\\
        \xmark & \cmark & \xmark & 35.4$\pm$1.0 & 48.5$\pm$1.4 & 30.9$\pm$1.0 & 43.8$\pm$1.4\\
        \xmark & \xmark & \cmark & 34.9$\pm$0.7 & 48.2$\pm$0.9 & 29.4$\pm$0.8 & 42.7$\pm$0.9\\
        \cmark & \cmark & \xmark & \textbf{35.8$\pm$0.9} & \textbf{49.0$\pm$1.4} &
                               \textbf{31.8$\pm$1.0} & \textbf{44.3$\pm$1.4}\\
      \bottomrule
    \end{tabular}}%
    
    \label{tab:model_component}
    \vspace{-12pt}
\end{table}

\begin{table}[t]
  \caption{Tiny\,ImageNet ($T\in\{10,20\}$): Performance impact of training-batch composition (\emph{Method}, $N$).}
  \centering
  \small
  \setlength{\tabcolsep}{4pt}
  \renewcommand{\arraystretch}{0.9}
  \scalebox{0.92}{%
    \begin{tabular}{@{}cc|cc|cc@{}}
      \toprule
      \multicolumn{2}{c|}{\textbf{Sampling}} &
      \multicolumn{2}{c|}{$T{=}10$} &
      \multicolumn{2}{c}{$T{=}20$} \\
      Method & $N$ &
      $A_{\text{last}}$(\%) & $A_{\text{inc}}$(\%) &
      $A_{\text{last}}$(\%) & $A_{\text{inc}}$(\%) \\
      \midrule
      Gaussian \cite{Magistri2024EFC}       &  64  & 34.5$\pm$1.1 & 47.9$\pm$1.6 & 28.4$\pm$1.2 & 42.1$\pm$1.6 \\
      Gaussian \cite{Magistri2024EFC}       & 128  & 34.0$\pm$1.1 & 47.6$\pm$1.5 & 27.9$\pm$1.1 & 41.8$\pm$1.3 \\
      Gaussian \cite{Magistri2024EFC}       & 256  & 33.3$\pm$1.4 & 47.8$\pm$1.6 & 27.2$\pm$1.6 & 41.4$\pm$1.9 \\
      Bi-interpolate \cite{shi2023prototype} &  64  & 32.8$\pm$1.1 & 47.2$\pm$1.3 & 26.7$\pm$1.0 & 41.0$\pm$1.1 \\
      CEOS           &  64  & \textbf{35.1$\pm$1.2} & \textbf{48.7$\pm$1.5} & \textbf{30.4$\pm$1.1} & \textbf{43.0$\pm$1.4} \\
      \bottomrule
    \end{tabular}}%
  \label{tab:batch_comp_t10_t20}
  \vspace{-16pt}
\end{table}

\vspace{-1mm}
\subsection{Ablation Study}
\vspace{-1mm}


\noindent \textbf{Ablation study on different components.} Table~\ref{tab:model_component} probes the effectiveness of the two components
proposed in this work—\emph{Constrained Expansive Over-Sampling} (CEOS) and the
\emph{Adaptive Class-Balanced} loss \(L_{\text{acb}}\).  
Because Focal loss~\cite{Lin:2020} is a popular remedy for
class imbalance, it is also evaluated for reference.


Starting from the plain baseline, \textbf{CEOS} alone provides a modest yet
consistent lift in both \(A_{\text{last}}\) and \(A_{\text{inc}}\) for the
10-task and 20-task splits.  
Activating only the \textbf{adaptive class-balanced loss}
\(L_{\text{acb}}\) yields a similar improvement, showing that time-aware
re-weighting is as effective as explicit minority over-sampling.  
Replacing standard cross-entropy with \textbf{Focal loss} offers only a marginal gain and still trails the performance achieved by \(L_{\text{acb}}\).

\noindent \textbf{Gaussian vs.\ Manifold Prototype Sampling.}
Table~\ref{tab:batch_comp_t10_t20} compares Gaussian sampling, our manifold oversampling (CEOS), and a bi\mbox{-}interpolate variant across $T{=}10$ and $T{=}20$. 
Holding the batch size fixed, adding \emph{more} Gaussian prototype samples does not help: at $T{=}20$, $A_{\text{last}}$ decreases monotonically as $N$ grows (28.4$\!\to\!$27.9$\!\to\!$27.2) with a slight decline in $A_{\text{inc}}$ (42.1$\!\to\!$41.8$\!\to\!$41.4); a similar pattern appears at $T{=}10$ (34.5$\!\to\!$34.0$\!\to\!$33.3). 
This indicates that the difficulty is not merely old–new imbalance but also \emph{how} rehearsal samples populate the space: Gaussian draws assume local spherical structure and tend to generate off\mbox{-}manifold or overlapping points that poorly match the evolving geometry. 
In contrast, introducing a small number of \emph{manifold} samples via CEOS ($N{=}64$) yields the best results in both regimes—\textbf{35.1}/\textbf{48.7} at $T{=}10$ and \textbf{30.4}/\textbf{43.0} at $T{=}20$—suggesting better boundary alignment and discrimination among older prototypes.

\noindent \textbf{Bi-interpolate vs.\ CEOS.}
We further include a \emph{bi\mbox{-}interpolate} baseline reimplementing PRAKA~\cite{shi2023prototype}, which synthesizes data by thresholded mixing of new features with old prototypes. 
Although superficially similar to CEOS, PRAKA randomly selects the old prototype to pair with; as illustrated in Figure~\ref{fig:abcd}, pairing with a farther prototype can push synthesized points into regions that overlap with neighborhoods of other (closer) prototypes, harming separability. 
Empirically, bi\mbox{-}interpolate underperforms CEOS (e.g., $T{=}20$: 26.7/41.0 vs.\ 30.4/43.0; $T{=}10$: 32.8/47.2 vs.\ 35.1/48.7), underscoring the importance of \emph{manifold\mbox{-}aware, boundary\mbox{-}aligned} oversampling over random prototype pairing.

\begin{figure}[t]
  \centering
  \includegraphics[width=.75\columnwidth]{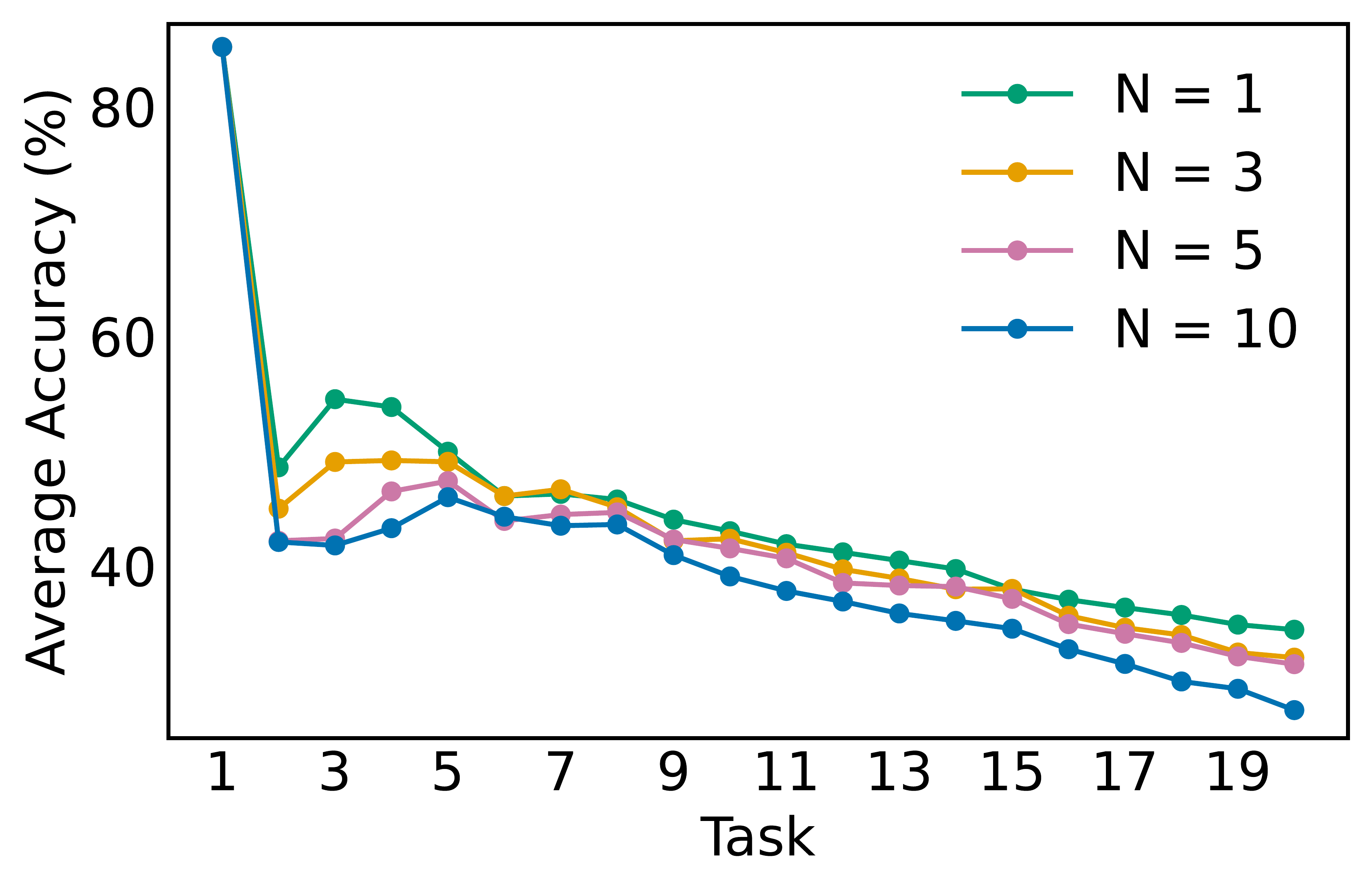}
  \caption{Tiny~ImageNet ($T{=}20$): Accuracy under different numbers of enemies used in CEOS.}
  \label{fig:enemy_num}
  \vspace{-12pt} 
\end{figure}

\noindent \textbf{Impact of the number of nearest enemies on CEOS.} Figure~\ref{fig:enemy_num} shows the performance of various numbers of nearest enemies selected for performing CEOS during prototype rehearsal. While in oversampling for static data the higher number of nearest enemies tends to yield better results \cite{dablain2023efficient}, in the analyzed EFCIL setting we observe the best performance when using 1 nearest enemy, especially on early and late tasks. This suggests that over-expanding the synthetic support region via multiple enemy interpolations may introduce ambiguous and overlapping features that lead to task confusion and degrade classifier performance. In contrast, using only the closest enemy ensures that synthetic samples remain tightly anchored to the class boundary while maintaining semantic coherence, resulting in more effective prototype reinforcement and better retention over time. These findings highlight the importance of controlled, minimally invasive augmentation in preserving the discriminative structure of old classes for prototype rehearsal in EFCIL. The best results arise when \textbf{CEOS} and \textbf{\(L_{\text{acb}}\)} are combined,
delivering the top scores on both \(A_{\text{last}}\) and \(A_{\text{inc}}\), indicating their synergy.





\newcommand{\subcap}[1]{\footnotesize\textbf{#1}}

\begin{figure}[t]
  \centering
  \setlength{\abovecaptionskip}{4pt}
  \setlength{\belowcaptionskip}{-6pt}

  \begin{minipage}[t]{0.49\columnwidth}
    \centering
    \includegraphics[width=\linewidth]{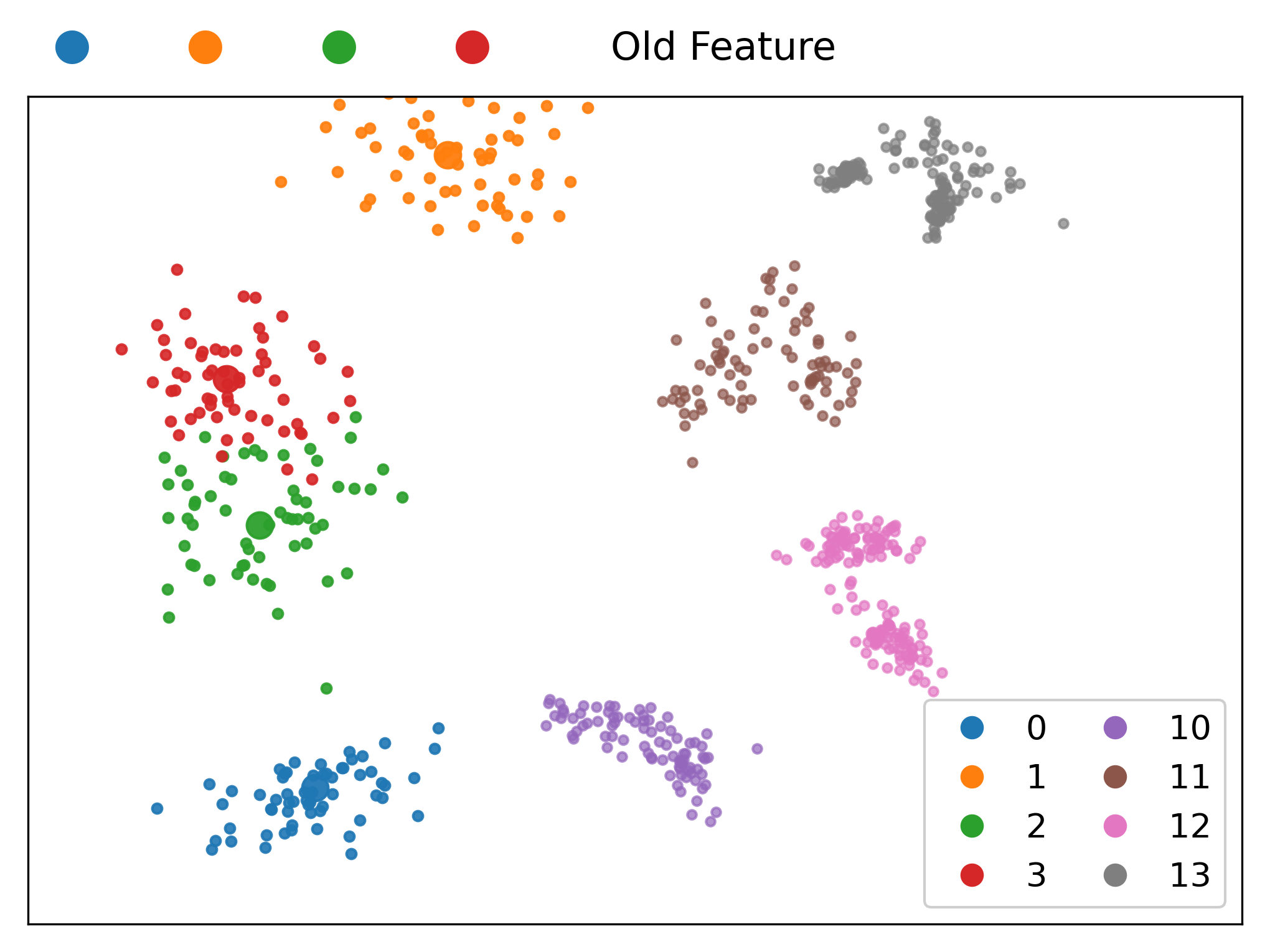}
    \\[-3pt]\subcap{(a)} Gaussian Sampling \cite{Magistri2024EFC}
  \end{minipage}\hfill
  \begin{minipage}[t]{0.49\columnwidth}
    \centering
    \includegraphics[width=\linewidth]{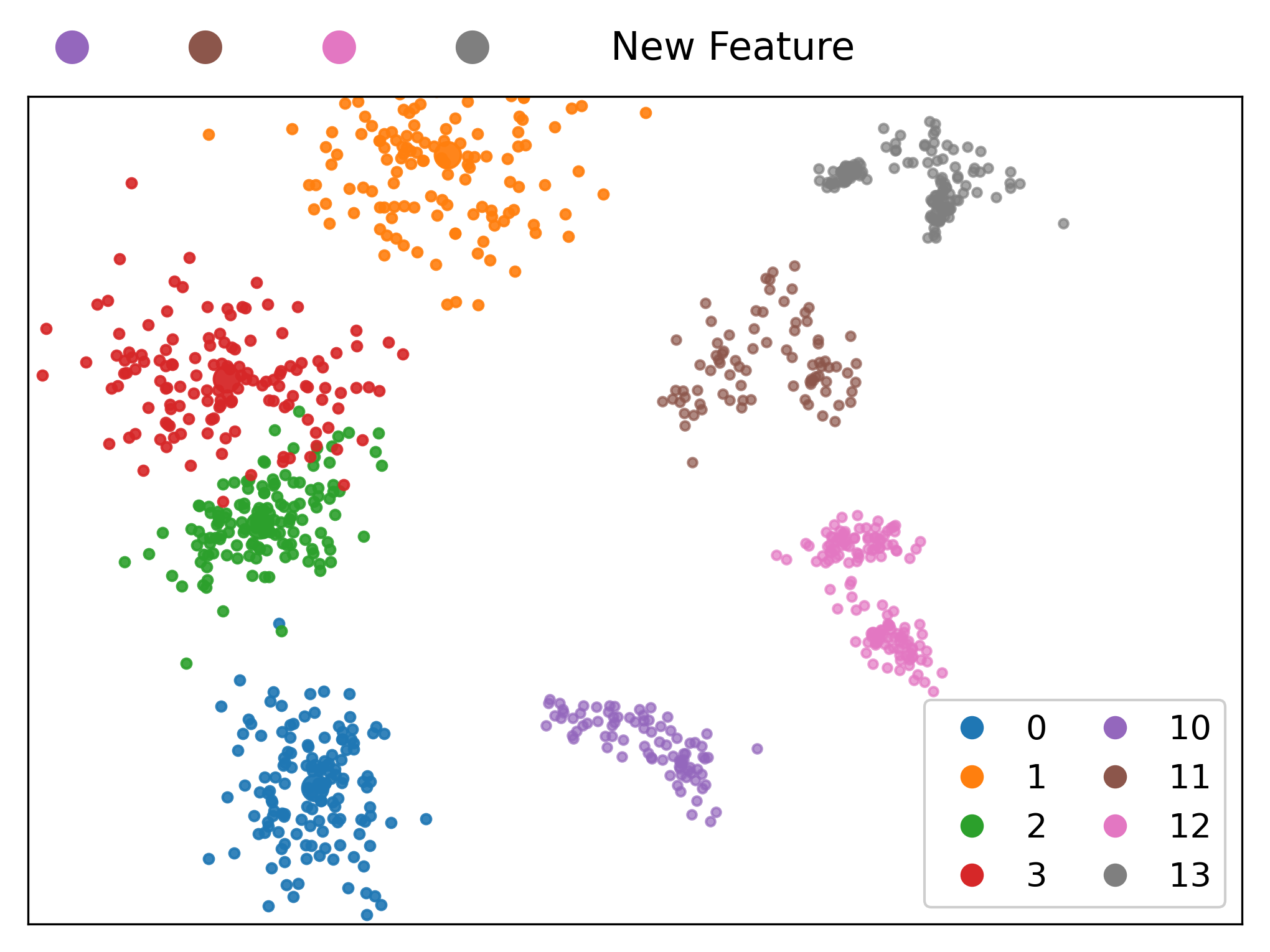}
    \\[-3pt]\subcap{(b)} Heavy Gaussian Sampling \cite{zhu2021pass}
  \end{minipage}

  \vspace{3pt} 

  \begin{minipage}[t]{0.49\columnwidth}
    \centering
    \includegraphics[width=\linewidth]{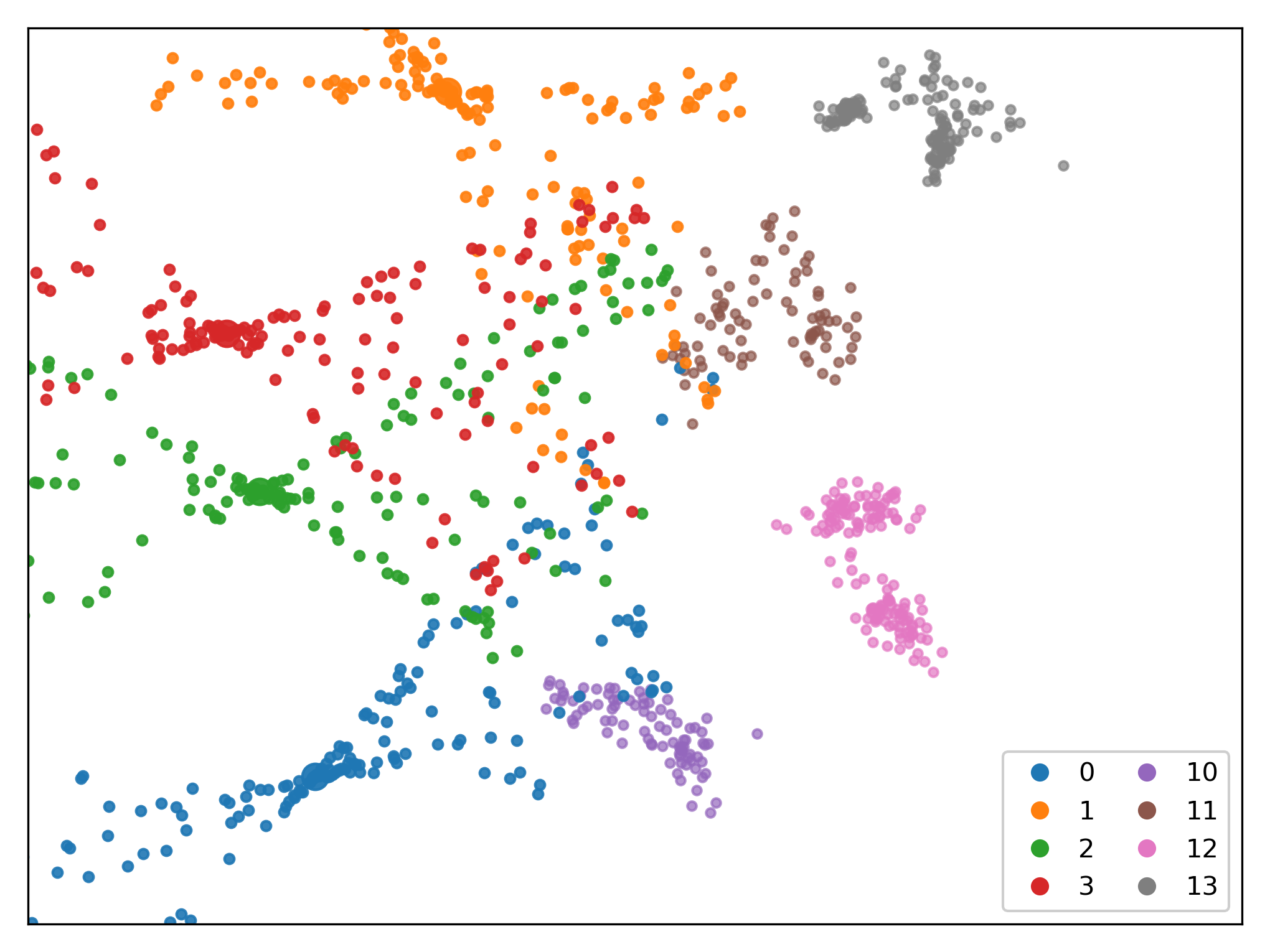}
    \\[-3pt]\subcap{(c)} PRAKA~\cite{shi2023prototype}
  \end{minipage}\hfill
  \begin{minipage}[t]{0.49\columnwidth}
    \centering
    \includegraphics[width=\linewidth]{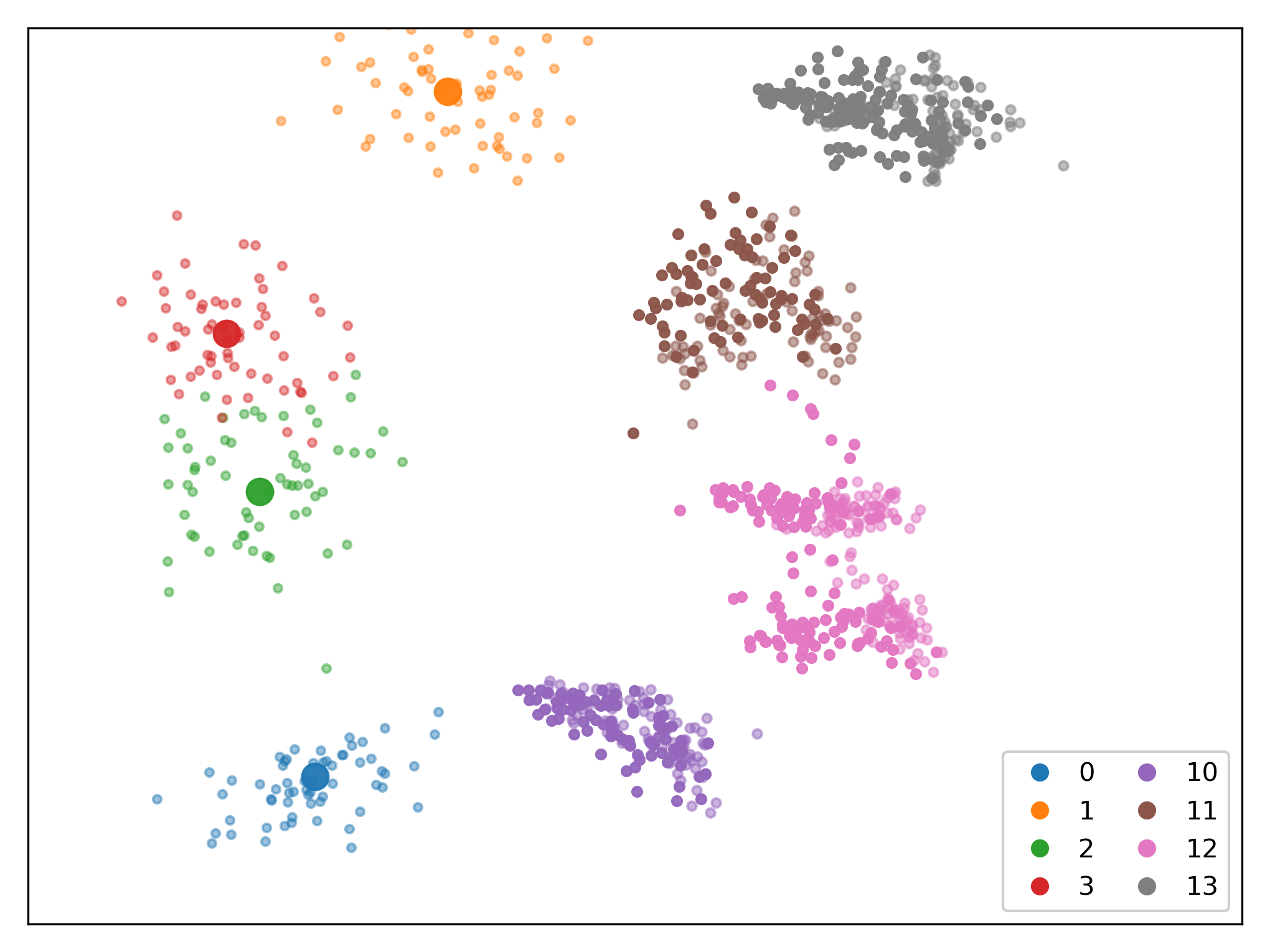}
    \\[-3pt]\subcap{(d)} CEOS
  \end{minipage}

  \caption{\textbf{Old vs.\ new features with different synthesis strategies.}
  (a) Gaussian Sampling. (b) Heavy Gaussian Sampling.
  (c) Random bidirectional interpolation.
  (d) proposed CEOS.}
  \label{fig:abcd}
  \vspace{-12pt}
\end{figure}

\begin{wrapfigure}{r}{0.47\columnwidth}
\vspace{-12pt}
  \centering
  \includegraphics[width=.45\columnwidth]{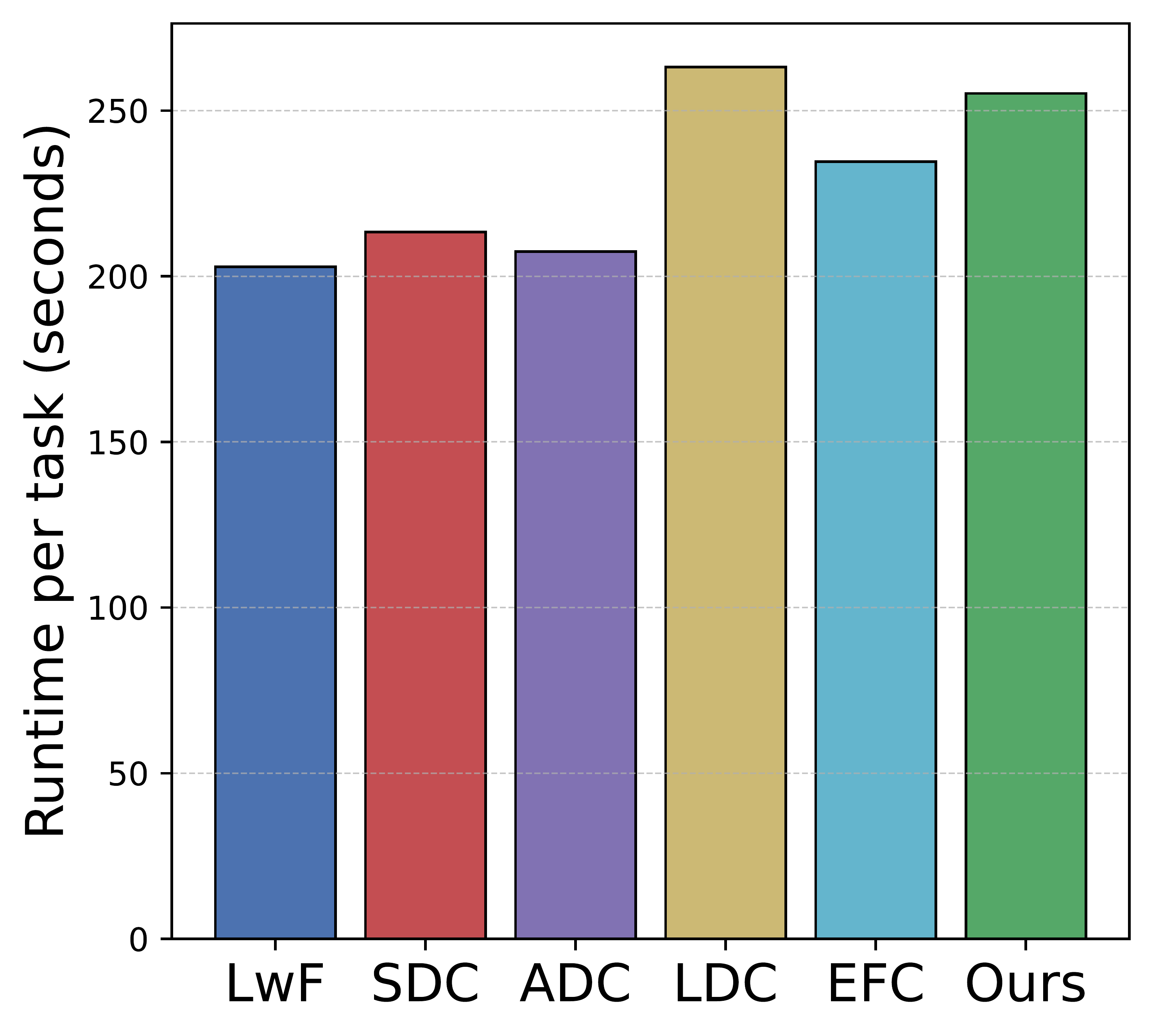}
  \caption{Average running time for 100 epochs per task on CIFAR-100 (10 splits, N = 128).}
  \label{fig:vanilla}
  \vspace{-12pt}  
\end{wrapfigure}

\noindent \textbf{Time Complexity Comparison.} We measured the training time of our method and the competitors using their original code on a workstation equipped with an NVIDIA RTX~6000 Ada GPU and a Xeon Gold 6448Y CPU. Each experiment was repeated five times; every run trained the model for 100 epochs on a \textsc{CIFAR-100} 10-task split with four data-loading workers and a batch size of 128. Our prototype-rehearsal strategy achieves highly competitive computational efficiency relative to drift-compensation methods. As shown in Figure~\ref{fig:vanilla}, LwF remains the fastest baseline (\(\approx\!200\) seconds per task), but prototype rehearsal---built on top of the lightweight EFC pipeline---adds only a minimal overhead and stays within the same computational envelope.

In contrast, drift-aware approaches such as SDC and ADC introduce additional forward- or statistics-tracking steps, resulting in modest slowdowns of about \(+5\%\) and \(+3\%\), respectively. LDC is the most computationally demanding, requiring 30 additional projector-training epochs and exceeding 260 seconds per task.

\vspace{-1mm}
\section{Conclusions, Limitations, and Future Works}\label{conclusion}
\vspace{-1mm}

\noindent \textbf{Conclusions.}  
We revisit prototype rehearsal in EFCIL and show that its standard design hides a persistent class imbalance caused by under-represented, drifting prototypes. Building on this analysis, we introduce a new strategy that expands prototype support through boundary-aware Constrained Expansive Over-Sampling and stabilizes optimization with an Adaptive Class-Balanced loss, achieving new state-of-the-art results for rehearsal-based approaches.

\noindent \textbf{Limitations.}  
Our method still assumes that prototype estimates remain sufficiently informative as the feature space drifts, and it relies on the presence of meaningful enemy neighbors for boundary-aware augmentation. In highly non-stationary or sparsely populated regions, interpolation may occasionally produce ambiguous samples.

\noindent \textbf{Future Work.}
We will relax these assumptions via uncertainty- and drift-aware prototype weighting and adaptive interpolation aligned with evolving manifolds, and extend our prototype rehearsal framework to transformer and multimodal continual models.

{
    \small
    \bibliographystyle{ieeenat_fullname}
    \bibliography{main}
}

\end{document}